\documentclass[letterpaper, 10 pt, conference]{ieeeconf}  % Comment this line out if you need a4paper

\IEEEoverridecommandlockouts                              % This command is only needed if 
\usepackage{cite}
\usepackage{amsmath,amssymb,amsfonts}
\usepackage{algorithmic}
\usepackage{graphicx}
\usepackage{caption}    % For customizing captions
\usepackage{subcaption} % For creating subfigures
\usepackage[percent]{overpic}
\usepackage{tikz}
\usepackage{pgfplots}
\usepackage{textcomp}
\usepackage{xcolor}
\usepackage{placeins}
\usepackage{cite}
\def\BibTeX{{\rm B\kern-.05em{\sc i\kern-.025em b}\kern-.08em
    T\kern-.1667em\lower.7ex\hbox{E}\kern-.125emX}}

\title{\LARGE \bf
Diffusion for Multi-Embodiment Grasping
}

\author{Roman Freiberg$^{1,*}$, Alexander Qualmann$^{1}$,Ngo Anh Vien$^{1}$ and  Gerhard Neumann$^{2}$%
\thanks{$^{1}$ Bosch Corporate Research, \tt\small \{roman.freiberg, alexander.qualmann, anhvien.ngo\}@bosch.com}
\thanks{$^{*}$Corresponding author}%
\thanks{$^{2}$Karlsruhe Institute of Technology {\tt\small gerhard.neumann@kit.edu}}%
}

\pgfplotsset{compat=1.18}

\begin{document}

\maketitle
\bstctlcite{IEEEexample:BSTcontrol} 
\thispagestyle{empty}
\pagestyle{empty}

%%%%%%%%%%%%%%%%%%%%%%%%%%%%%%%%%%%%%%%%%%%%%%%%%%%%%%%%%%%%%%%%%%%%%%%%%%%%%%%%
\begin{abstract}
 Grasping is a fundamental skill in robotics with diverse applications across medical,
industrial, and domestic domains. However, current approaches for predicting valid grasps are
often tailored to specific grippers, limiting their applicability when gripper designs change.
To address this limitation, we explore the transfer of grasping strategies between various gripper
designs, enabling the use of data from diverse sources. In this work, we present an approach based
on equivariant diffusion that facilitates gripper-agnostic encoding of scenes containing graspable
objects and gripper-aware decoding of grasp poses by integrating gripper geometry into the model.
We also develop a dataset generation framework that produces cluttered scenes with variable-sized
object heaps, improving the training of grasp synthesis methods. Experimental evaluation on diverse
object datasets demonstrates the generalizability of our approach across gripper architectures,
ranging from simple parallel-jaw grippers to humanoid hands, outperforming both single-gripper
and multi-gripper state-of-the-art methods.

\end{abstract}

%%%%%%%%%%%%%%%%%%%%%%%%%%%%%%%%%%%%%%%%%%%%%%%%%%%%%%%%%%%%%%%%%%%%%%%%%%%%%%%%
\section{Introduction}  
Recent advancements in imitation learning and reinforcement learning have significantly enhanced data
efficiency and performance in robotic grasping methods. Notably, the integration of diffusion
models \cite{janner2022diffuser, urain2023se, ryu2023diffusion} has shown promising results
in handling multimodal grasping data. However, current grasping
methods are often designed for task-specific end-effectors and exhibit limited
transferability to
alternative gripper architectures. These methods are tailored to predefined hardware configurations,
assuming fixed characteristics such as degrees of freedom (DoF), gripper geometry, and the physical
contact mechanics of the end-effector. Transferability is crucial for developing generalizable grasp
detection methods that can work across various robot systems with different gripper designs.
It requires solutions that are resilient to specific hyperparameters and hidden biases, allowing
them to incorporate data from diverse sources \cite{li2023gendexgrasp, attarian2023geometry}. Incorporating gripper-agnostic design into architectures
could pave the way for future foundation models in robotic grasping \cite{bousmalis2023robocat, padalkar2023open}.
Our contribution is twofold. First, we propose an approach based on an $\textrm{SE}(3)$-equivariant
architecture \cite{ryu2023diffusion, liao2022equiformer} that demonstrates state-of-the-art performance in both
single-gripper settings and multi-gripper benchmarks, highlighting the effectiveness of integrating gripper
geometry with feature adaptation across diverse gripper configurations.
We directly put the diffusion process always relative to the gripper frame,
which simplifies the equivariance conditions and improves performance.
Second, we provide experimental validation of our approach on diverse object datasets
\cite{calli2017yale, downs2022google}. Finally, we plan to open-source our multi-gripper grasp scene
generation framework, which integrates eight gripper types and over 1,000 objects. Our findings strongly 
suggest that training on cluttered heaps is beneficial for grasp synthesis methods; as such,
we include a pipeline for variable-size object heap generation in both tabletop and bin picking settings.

\begin{figure}
  \centering
  \begin{overpic}[width=0.9\linewidth]{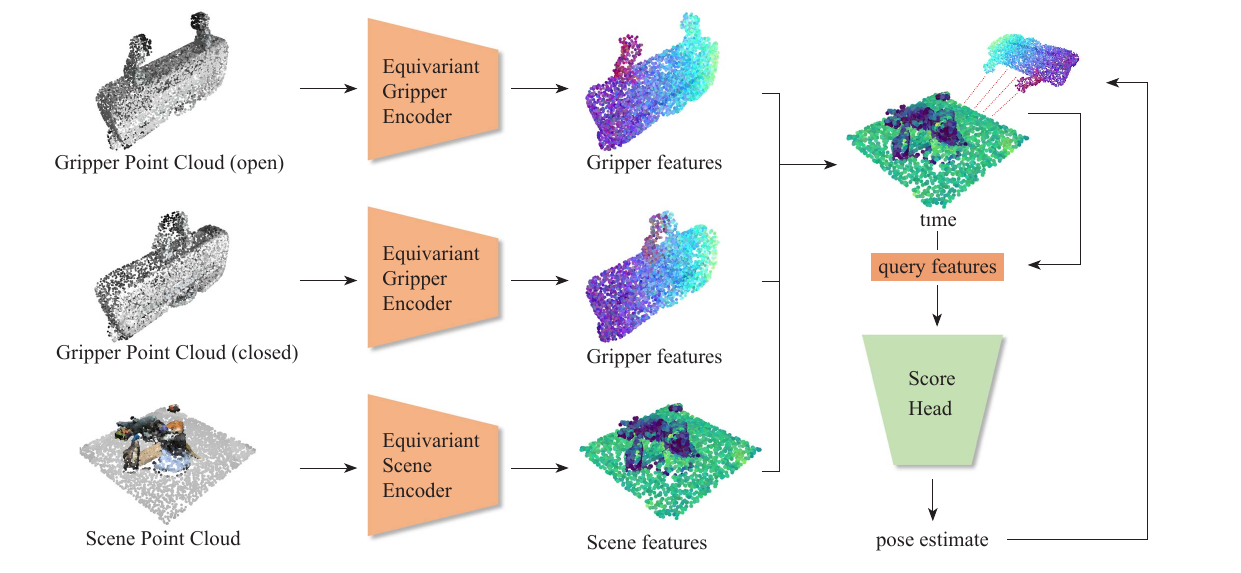}
    \put(20, -4){(a) Encoding}
    \put(67, -4){(b) Diffusion}
  \end{overpic}
  \vspace{5pt}
    \caption{\textbf{Overview of architecture}. \textbf{(a)} Point cloud scans of scene with objects and 
    gripper in open and closed configurations are equivariantly encoded to a point cloud feature space. 
    \textbf{(b)} Diffusion process computes pre-grasp pose for given gripper and scene
    using feature query for current pose estimate.}
  \label{fig:overview}
\end{figure}

\section{Related Work}

\paragraph{Grasp Detection Methods and Datasets} Data-driven approaches have significantly advanced
single-gripper grasping, covering both open-loop and closed-loop strategies \cite{fang2020graspnet, schillinger2023model, mahler2018dex, song2020grasping, xu2021adagrasp, sundermeyer2021contact, fang2023anygrasp, acronym2020}. By
incorporating visual features and gripper-specific priors, these methods have been tailored
to end-effectors like parallel-jaw or vacuum grippers. Such approaches effectively leverage
fixed hardware configurations, facilitating learning-based strategies. However, their specificity
often limits their applicability to other gripper designs. This limitation has spurred interest
in multi-embodiment agents and robotics foundation models \cite{bousmalis2023robocat, reed2022generalist, padalkar2023open}.
Several datasets have been introduced for robotic grasping, each focusing on single gripper types.
The Cornell Grasp Dataset \cite{5980145} provides annotated RGB-D images for grasp detection with
a specific gripper. GraspNet-1Billion \cite{fang2020graspnet} and EGAD! \cite{morrison2020egad}
offer extensive grasp poses, but primarily for parallel-jaw grippers. Other works, including
Mahler et al.\cite{mahler2017dex} and Eppner et al.\ \cite{acronym2020}, present large-scale grasp
datasets but focus on specific grippers like the Franka Panda. This lack of datasets encompassing
multiple gripper types across diverse objects highlights the need for our custom dataset.

\paragraph{Generalist Agents}
Generalist methods aim to minimize inductive biases by treating grasping as a secondary objective within a range of
diverse tasks \cite{bousmalis2023robocat, team2024octo}. While these "one-size-fits-all" representations are
versatile and can handle various tasks, they may not emphasize grasp-specific details such as gripper
configurations. In contrast, generalizable grasp representations focus on developing models that can transfer
across different grasp distributions. A notable example of this approach is the use of contact maps,
which highlight successful grasp regions on objects \cite{attarian2023geometry, li2023gendexgrasp}.
Gripper-specific encodings directly predict stable grasps by learning gripper-specific information.
However, methods that focus solely on top-grasp selection, like AdaGrasp \cite{xu2021adagrasp}, may limit broader
applicability. Our work aligns with this category, as we believe it fosters models that
are better suited for gripper generalization.

\paragraph{Diffusion for Grasp Synthesis}
Diffusion methods have revolutionized distribution modeling \cite{ho2020denoising, corso2023diffdock, song2020score, de2022riemannian, ryu2023diffusion}, particularly for the multimodal
distributions present in grasp data \cite{simeonov2023shelving, urain2023se, ryu2023diffusion}.
Advances in manifold modeling, such as on $\textrm{SE}(3)$,
have further enabled learning of grasp poses \cite{ryu2023diffusion, leach2022denoising}.
Additionally, zero-shot domain transfer
has become an essential technique for bridging the gap between different data distributions,
with notable examples including CycleGAN \cite{zhu2017unpaired} and the Mirage project \cite{chen2024mirage}. While these methods are effective,
we avoid relying on a single physical gripper model as the sole reference to prevent potential biases.

\paragraph{Equivariant Methods} There has been a growing interest in exploiting data symmetries
in grasping tasks. Since the pioneering work of Cohen and Welling \cite{pmlr-v48-cohenc16},
which introduced the concept of discrete group convolutions in CNN-based architectures, there has
been substantial ongoing research in this field. Subsequent investigations
\cite{worrall2017harmonic, bekkers2020bspline, weiler2021coordinate} have expanded the methodology
by utilizing steerable basis functions, enabling the representation of continuous groups such as $\textrm{E}(N)$
on Riemannian manifolds.
Compared to traditional approaches, equivariant designs typically exhibit superior generalization and
data efficiency \cite{bekkerroto}.
Equivariance, particularly with respect to rotations and translations, offers significant gains
in data efficiency, making it highly beneficial for robotic applications
\cite{liao2022equiformer, ryu2022equivariant, yang2024equibot}. Additionally,
graph-based networks have proven effective in modeling structurally complex problems across
varying data scales \cite{scarselli2008graph, dwivedi2020generalization, li2019deepgcns}. Their
adaptability in handling diverse data inputs has made them increasingly popular in robotics, particularly
for point cloud and context modeling \cite{lou2022learning, huang2023defgraspnets, chun2023local, simeonov2022neural}.

\section{Problem Formulation and Preliminaries}
\label{sec:terms}
We assume a given set of grippers $\{ G_1, \ldots, G_n\}$, each associated with internal joint
configurations $\{J_1, \ldots, J_n\}$ with variable DoF. Prominent examples include the gripper
width for parallel-jaw architectures, such as the Franka Gripper, and the 20 DoFs in the
finger joints of the Shadow Hand. In this work, we predominantly focus on the sub-problem with fixed configurations
while discussing possible extensions of our approach to the original problem statement. As such,
each gripper is described by a pre-grasp and the corresponding target grasp configuration. Different
gripper configurations are treated as distinct gripper types. This is exemplified in our use of the Shadow
Hand in various gripper configurations, including two-finger, three-finger, and full-hand grasps,
treating each configuration as a distinct gripper type. 
We denote $o_{G_i}$ as the observation of the gripper in its local coordinate frame, consisting
of open and closed point clouds rendered from the open and closed gripper geometries, respectively.
The target policy distribution is defined by $p(r \mid o_{G_i}, o_s)$, where
$r \in \textrm{SE}(3)$ represents the target pre-grasp pose, and $o_s$ is the observation of
the scene $s$ in the world coordinate frame containing at least one graspable object. Potential obstacles are
treated as part of the scene.
\subsection{Diffusion for Generative Modeling}
Diffusion models have proven to be a viable method for modeling high-dimensional
multimodal distributions. Standard diffusion models on $ \mathbb{R}^d $ are based on the
Ornstein-Uhlenbeck process \cite{baldi2017stochastic}, described by
\[
d\mathbf{x} = -{\beta(t)}\mathbf{x} \, dt + \sqrt{2{\beta}(t)} \, d\mathbf{B}, \quad \mathbf{x}(0) \sim p_0,
\]
where $\mathbf{x}(t)$ is an adaptable process on $ \mathbb{R}^d $, and $\mathbf{B}(t)$
is the Brownian motion. As time $t$ approaches infinity, this distribution converges to
the normal distribution with zero mean and $I_d$ covariance,
where $I_d$ is the d-dimensional identity matrix.
Here, $\mathbf{x}(0)$ represents the process origin at initial time. The reverse process, given by
\begin{equation}
\label{eq:rev-dif}
d\mathbf{x} = -\beta(t)\left(\mathbf{x} + 2\nabla \log(p_t(\mathbf{x}))\right)\, dt + \sqrt{2\beta(t)} \, d\mathbf{B},
\end{equation}
allows the computation of the backward path during inference by approximating the score
$\nabla \log(p_t({\mathbf{x}}))$ of the process through optimizing the corresponding
MSE loss formulation \cite{ho2020denoising}.

\subsection{Diffusion on \textrm{SE}(3)}
For compact manifolds such as $\textrm{SO(3)}$, De Bortoli et al. \cite{de2022riemannian} proposed the corresponding
process $d\mathbf{x} = d\mathbf{B}^{\textrm{SO}(3)}$,
to fully exploit the data structure, where $\mathbf{B}_t^{\textrm{SO}(3)}$ represents Brownian
motion on the $\textrm{SO(3)}$ manifold. The $\textrm{SE(3)}$ manifold, which combines $\textrm{SO(3)}$
for rotation and $\mathbb{R}^3$ for translation, requires a combined process,
$p(r_t \mid r_0) = \mathbf{B}_t^{\mathrm{SE}(3)}(r_0^{-1}r_t)$, where $r_t, r_0 \in \textrm{SE(3)}$.
A pose  $r = (\omega, p)$ can be decomposed into its rotational and translational parts, where $\omega$ is the rotation angle in the axis-angle parametrization, $p$ is the translation.  Similarly, the Brownian motion on $\textrm{SE(3)}$ can be decomposed into 
\[
\mathbf{B}_t^{\textrm{SE}(3)}(\omega, p) =
\mathcal{N}(p; p, t I_d)\mathcal{IG}_{SO(3)}(\omega, t).
\]
Here, $\mathcal{IG}_{\textrm{SO}(3)}$ is the isotropic Gaussian on $\textrm{SO(3)}$  defined
by the relation
\[ 
\mathcal{IG}_{\textrm{SO}(3)}(\omega, t) \propto \sum_{l=0}^{\infty}(2l + 1)e^{-t l(l+1) / 2} \frac{\sin((l+1/2)\omega)}{\sin(\omega/2)}.
\]
This formulation allows for the numerical simulation
of the forward path, following the sampling procedure presented in the work of Leach et al. \cite{leach2022denoising}
and Ryu et al. \cite{ryu2023diffusion}.

\subsection{Equivariance on \textrm{SE(3)}}
A function $f: X \to Y$, between two vector spaces $X, Y$, is said to be equivariant with respect
to the group action $r \in \textrm{SE(3)}$ if it satisfies the condition
\begin{equation}
\label{eq:eq}
    f(D_X(r)x) = D_Y(r)f(x),
\end{equation}
where $x \in X$, $y \in Y$, and $D_{X/Y}(r)$ are transformation matrices parameterized by
$r$ representing the group action as a matrix in the respective domain.
% Architectures exemplifying this relationship are presented
% in the work of Liao et al. \cite{liao2022equiformer} and Ryu et al. \cite{ryu2023diffusion}. 

\begin{figure*}
  \centering
  \begin{overpic}[width=0.95\linewidth]{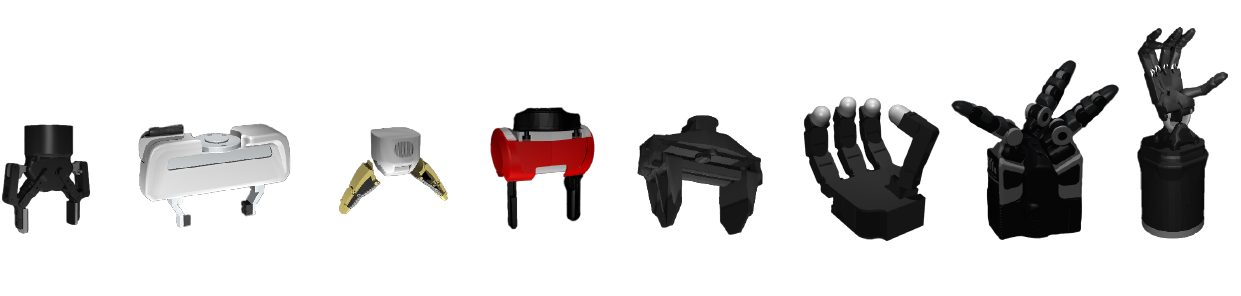}
    \put(3, 2){(a)}
    \put(17, 2){(b)}
    \put(31, 2){(c)}
    \put(43, 2){(d)}
    \put(57, 2){(e)}
    \put(70, 2){(f)}
    \put(83, 2){(g)}
    \put(94, 2){(h)}
  \end{overpic}
    \caption{\textbf{Overview of used grippers}. \textbf{(a)} Robotiq 2F-85, \textbf{(b)} Franka Emika Gripper,
    \textbf{(c)} Google Bot Gripper, \textbf{(d)} Rethink Gripper, \textbf{(e)} ViperX 300s Gripper, \textbf{(f)} Allegro 
    \textbf{(g)} Shadow DEX-EE Hand \textbf{(h)} Shadow Hand
    }
  \label{fig:used-gripper}
\end{figure*}
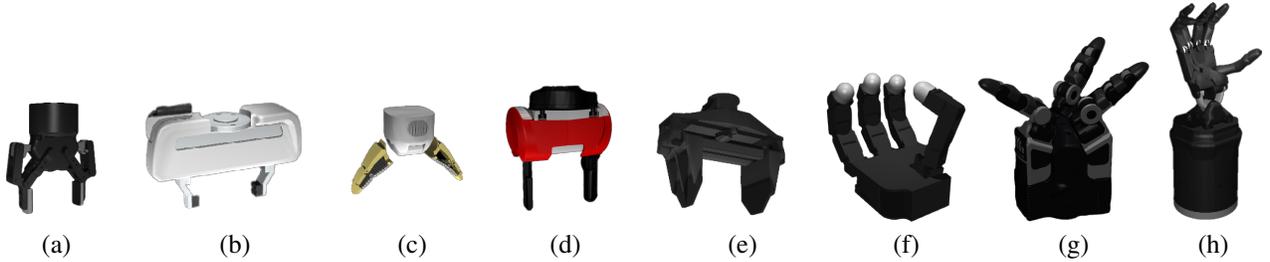

\section{Multi-Embodiment Grasping}
In order to achieve multi-embodiment grasping, we first need to create multi-gripper datasets for
training our grasp synthesis models. We use eight gripper types and generate grasps for single
objects as well as for cluttered scenes. In terms of grasp synthesis, we propose an architecture based
on equivariant diffusion for grasp prediction \cite{ryu2023diffusion}. Here, we introduce several
changes to this architecture to perform efficient multi-embodiment grasp synthesis. 

\subsection{Dataset Generation}

\begin{figure}
\centering
\begin{tikzpicture}
\begin{axis}[
   width=0.4\textwidth,
   height=5cm,
   ybar,
   ymin=0, ymax=300,
   symbolic x coords={1,2,3,4,5,6,7,8,9,10},
   xtick=data,
   x tick label style={rotate=45,anchor=east,font=\small},
   ylabel={Number of objects},
   ylabel style={font=\small},
   xlabel={Gripper able to grasp an object},
   xlabel style={font=\small},
   title={Count of Grippers Able to Grasp an Object},
   title style={font=\bfseries},
]
\addplot coordinates {
   (1,27) (2,34) (3,39) (4,67) (5,73) (6,84) (7,94) (8,146) (9,260) (10,200)
};
\end{axis}
\end{tikzpicture}
\vspace{5pt}
\caption{\textbf{Count of Grippers Able to Grasp an Object}. Grasp generation uses an antipodal sampling strategy for all gripper types, while preserving some gripper-specific properties.}
\label{fig:count}
\end{figure}
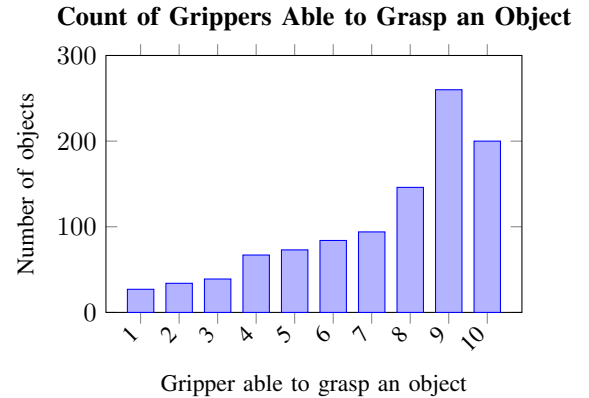

\paragraph{Grasp Synthesis}.
For our grasp synthesis procedure, we require a highly parallelizable
pipeline at any stage to manage computational complexity. To meet this requirement and avoid intensive manual
tuning, we opted for a single antipodal sampling strategy with individually aligned approaches and antipodal
directions for each gripper. Although this sampling strategy excludes grasp categories as
it does not fully exploit the gripper morphology, our generation method still captures gripper-specific
data, as summarized in Figure \ref{fig:count}. Thus, a gripper-agnostic transfer of grasps does not provide
a sufficient method to solve the original problem statement in our dataset.

\begin{figure}
    \centering
    \begin{subfigure}[b]{0.18\textwidth}
        \vspace{5pt}
        \raisebox{-4pt}{\includegraphics[width=\textwidth]{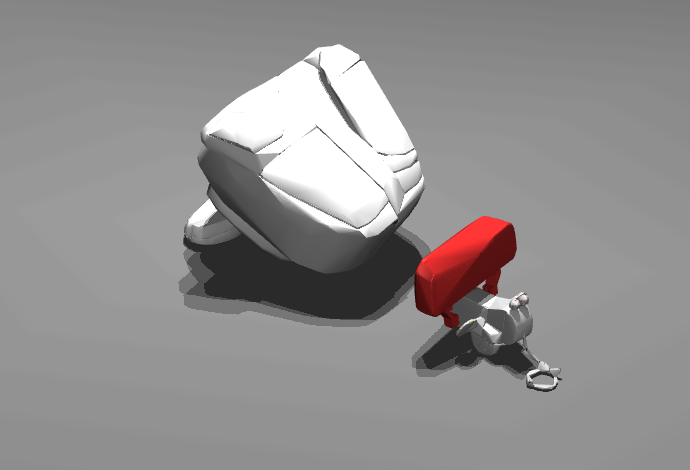}}
        \caption{Franka Gripper}
        \label{fig:image1}
    \end{subfigure}
    \begin{subfigure}[b]{0.18\textwidth}
        \vspace{5pt}
        \raisebox{-4pt}{\includegraphics[width=\textwidth]{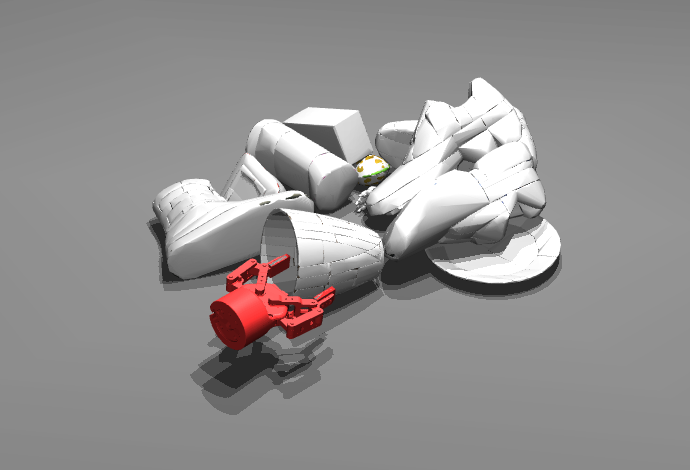}}
        \caption{Robotiq 2F-85}
        \label{fig:image2}
    \end{subfigure}
    \begin{subfigure}[b]{0.18\textwidth}
        \includegraphics[width=\textwidth]{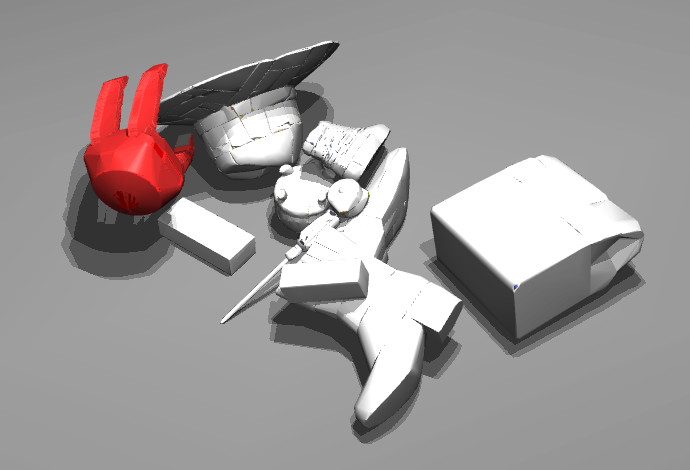}
        \caption{Robotiq 3-Finger}
        \label{fig:image3}
    \end{subfigure}
    % Fourth subfigure: Shadow Hand
    \begin{subfigure}[b]{0.18\textwidth}
        \includegraphics[width=\textwidth]{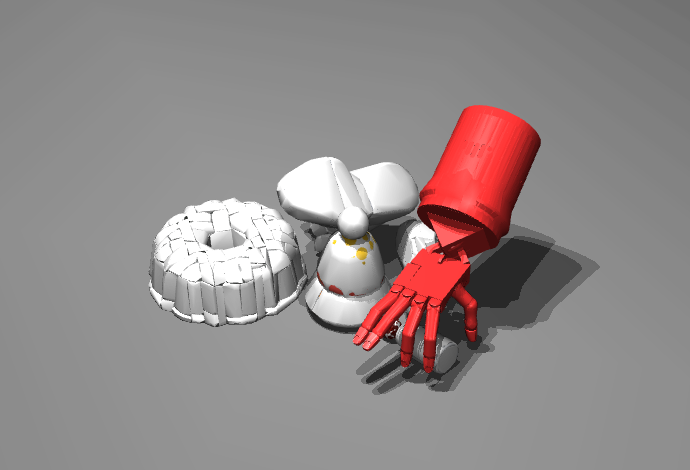}
        \caption{Shadow Hand}
        \label{fig:image4}
    \end{subfigure}
    \caption{\textbf{Collision models for grasp scenes}. Rendering of collision models in a grasping scene for various gripper types. Each subfigure illustrates an example of a generated grasp pose specific to the gripper type depicted.}
    \label{fig:collision}
\end{figure}

We used MuJoCo simulations \cite{todorov2012mujoco} and objects from the YCB \cite{calli2017yale} and Google Scanned Objects datasets \cite{downs2022google}. Due to the requirements
of MuJoCo for convex collision models, we preprocessed objects with V-HACD \cite{mamou2016volumetric} to approximate collision
surfaces while preserving simulation performance. Figure \ref{fig:collision} shows example renderings of collision models
of grasping scenes for several gripper types. After removing multi-part objects, the dataset comprised
over 1,000 objects with diverse sizes and categories. In some cases, objects were larger than the grippers,
necessitating precise placement for a successful grasp.
We used a variety of gripper models sourced from Google DeepMind’s Menagerie \cite{menagerie2022github}
and robosuite \cite{robosuite2020}, as depicted in Figure \ref{fig:used-gripper}. Our gripper set contains five two-finger grippers, one three-finger gripper, one four-finger gripper,
and three Shadow Hand configurations. Grasp generation
occurred in a gravity-free environment with shaking motions to determine stability, following the approach of Eppner et. al\ \cite{acronym2020}. In general, we
simulated 5,000 successful grasps per object-gripper pair. We allowed for fewer than 5,000 successful
grasps per gripper-object pair where the yield was below a threshold, defined by an
estimated time of arrival computed from prior statistics, making grasp synthesis computationally
costly.
All computations were performed on a high-performance
cluster running 1,500 CPU cores in parallel.

\paragraph{Clutter Scene Generation}
For grasp scene generation, we assembled heaps of objects, consisting of one to twelve items,
dropped onto a flat table.
Thus, our generation pipeline allows for variable complex heap generation.
Additionally, we developed a scene variant simulating a bin picking environment with randomly
generated variable sizes and colors for the bin to mitigate sim-to-real complications for our real-world
bin picking experiments in Section \ref{sec:results}. All environments introduced light and ground color 
randomizations to reduce overfitting on specific attributes, thus making the transfer to real
environments more robust. After a scene has settled into a stable configuration, previously
generated valid grasps were transformed to the object's pose. All grasps were reevaluated and
marked as stable when a gripper could lift an object while maintaining contact throughout the
lift operation. Grasps that resulted in a collision in the initial
pre-grasp pose were removed upfront.
In total, we synthesized and evaluated over 750 million grasps and distilled 2.5 million valid grasps from these
simulations across 450 simulation scenes per gripper variant for training.
Each sample includes an associated gripper and scene point cloud, sourced from dense and sparse
scans. Dense scans are obtained from ten color images containing depth information, while sparse
scans are obtained from a single image.
For future extensions, our framework supports the integration of all MuJoCo-compatible grippers \cite{menagerie2022github}, diverse object datasets, and various grasp sampling strategies.

\subsection{Architecture}

In the original architecture of Diffusion-EDFs \cite{ryu2023diffusion}, the equivariance
condition is expressed as
\begin{equation}
\label{eq:c1}
p(\Delta_r  r \mid o_{G_i}, \Delta_r  o_s) = p(r \mid o_{G_i}, o_s),
\end{equation}
where $\Delta_r \in \mathrm{SE}(3)$ represents an arbitrary transformation. An equivariant model must preserve this condition throughout the computational pipeline.
The authors of Diffusion-EDFs proposed a U-Net hierarchical encoding of the scene's point cloud.
At each layer, they downsample the point cloud using the farthest-point sampling (FPS) algorithm to
capture higher-level features. The equivariant descriptor field (EDF) at each stage
comprises points with associated features
composed of irreducible representations up to angular momentum type $l = 2$ \cite{e3nn_paper},
which can be transformed equivariantly under $\mathrm{SO}(3)$ actions.
In the decoder stage, a shallower network is used to query information at specified positions, dependent on 
the current diffusion time, in the hierarchical EDF. For the pick task,
the authors used predefined static query positions. During the diffusion process, the descriptor features
are transformed under the rotation group action using the corresponding Wigner D-matrices
for each irreducible representation. Finally, the score head employs the queried features to approximate
the score of the reverse diffusion process from Equation \ref{eq:rev-dif}, including the rotational component.

In order to make this architecture usable for multi-embodiment grasp synthesis,
we had to introduce the following adaptations to the Diffusion-EDF \cite{ryu2023diffusion} architecture.
\paragraph{Encoding the Gripper Geometry} As shown in Figure \ref{fig:overview}, we exploit the query
mechanism to incorporate the open and closed gripper geometries, conditioning the diffusion process on
end-effector-specific features. The FPS downsampled gripper point clouds
 encode gripper-specific features using a shallow U‑Net. These positions query features at
points transformed by the diffusion process. For each layer, the query and EDF point cloud,
along with the time embeddings, are passed to the query mechanism, which computes relative position
and angular edge encodings between both point clouds. These encodings, along with all components,
are passed to an equiformer layer that outputs layer-wise field
values. Summing over all layers produces the query features, which are fed to the score head.

\paragraph{Using the Gripper-Frame as Reference-Frame}
Originally, the authors specified the grasp equivariance condition expressed
as
\[
s( r \Delta^{-1}_r \mid \Delta_r o_{G_i}, o_s) = \left[\textrm{Ad}_{\Delta_r}\right]^{-T}s(r \mid
o_{G_i}, o_s),
\]
where $s( r \mid o_{G_i}, o_s)$ is the score function and 
$\textrm{Ad}_{\Delta_r}$ is an adjoint representation of the SE(3) transformation $\Delta_r$ 
\cite{ryu2023diffusion}. The adjoint representation is defined as
\[
\text{Ad}_{r=(p, \omega)} = \begin{bmatrix}
\omega & [p]^\wedge \omega \\
\mathbf{0} & \omega
\end{bmatrix},
\]
where $[p]^\wedge$ is the skew-symmetric matrix of $p$. This relation emerges naturally
from the computation of the right equivariant Lie-derivative of the score function
(see \textrm{Proposition 1.} in the work of Ryu et. al\ \cite{ryu2023diffusion}).
Achieving this condition as well as the equivariance condition from Equation \ref{eq:c1} 
requires the introduction of a reference frame selection mechanism.
However, in our case, we do not require this grasp equivariance condition because the
diffusion process always takes place in the gripper's reference frame.
Since we are interested in grasp synthesis relative to a fixed gripper frame, enforcing reference
frame independence is unnecessary. By not imposing this condition, we simplify the model,
which significantly facilitates training and increases performance.

\paragraph{Equivariant FiLM Layers for Multi-Resolution Diffusion}
The original model was split into two parts: one handling the low-resolution diffusion and
the other the high-resolution diffusion process.
We developed a model variant introducing an equivariant version of the FiLM \cite{perez2018film}
layer, commonly used in diffusion architectures \cite{3d_diffuser_actor}. Specifically, these
layers modulate the encoded equivariant features element-wise for each angular momentum type $l$ at
each timestep by applying scalar shifts and biases to each feature channel corresponding to the
irreducible representations, ensuring that the modulation respects the $\mathrm{SO}(3)$ equivariance.
The modulation parameters are generated from the time embeddings through learnable mappings and
are applied via equivariant tensor products. This design allows the model to account for
distribution shifts during the diffusion process. Implementing this variant enabled us to
train a single unified model, reducing the total training time by up to 35 percent.
Furthermore, our data provided grasp-object relations; thus, we trained on batches of grasps
sampled from diverse objects (biased sampling) to mitigate distribution bias toward objects with
more successful grasps.

% \begin{figure}
%     \centering
%     \begin{subfigure}[b]{0.18\textwidth}
%         \includegraphics[width=\textwidth]{corl/images/col_scenes/panda.png}
%         \caption{Franka Gripper}
%         \label{fig:image1}
%     \end{subfigure}
%     \begin{subfigure}[b]{0.18\textwidth}
%         \includegraphics[width=\textwidth]{corl/images/col_scenes/robo_2f.png}
%         \caption{Robotiq 2F-85}
%         \label{fig:image2}
%     \end{subfigure}
%     \begin{subfigure}[b]{0.18\textwidth}
%         \includegraphics[width=\textwidth]{corl/images/col_scenes/robotiq_s.png}
%         \caption{Robotiq 3-Finger}
%         \label{fig:image3}
%     \end{subfigure}
%     \begin{subfigure}[b]{0.18\textwidth}
%         \includegraphics[width=\textwidth]{corl/images/col_scenes/shadow_two.png}
%         \caption{Shadow Hand}
%         \label{fig:image4}
%     \end{subfigure}
%     \caption{\textbf{Collision models for grasp scenes}. Rendering of collision models in a
%     grasping scene for various gripper types. Each
%     subfigure illustrates an example of a generated grasp pose specific to the gripper type depicted.}
%     \label{fig:collision}
% \end{figure}

% TODO: Could you summarize how many grasps we have here just to show the scale of the dataset? (done)
% 

\section{Experimental Evaluations}
\label{sec:results}

We evaluate our approach to address several key questions, including whether our method sacrifices
single-gripper performance, if it successfully captures the multimodalities present in gripper-specific grasps,
and how performance is impacted in zero-shot and few-shot tasks.
Additionally, we examine whether a model trained on our data is applicable to a real robotic
bin picking system.

\begin{table}[h]
\centering
\begin{tabular}{l|c}
\textbf{Method} & \textbf{Success Rate (\%)} \\ \hline
Non-equivariant U-Net & 52.3 \\
Contact-GraspNet & 54.2 \\
Diffusion-EDF & 63.3 \\
Ours (only Franka gripper) & \textbf{94.5} \\
\
\end{tabular}
\caption{\textbf{Single-Gripper Benchmark Results}.
Average grasp success rates for Diffusion-EDF \cite{ryu2023diffusion},
Contact-GraspNet \cite{sundermeyer2021contact} (CGN) with original weights, a
non-equivariant model, and our method. Except for CGN, all methods were trained on 2,500
scenes using only the Franka gripper.
Results show that our approach does not compromise on single-gripper performance.}
\label{table:cgn}
\end{table}

\paragraph{Evaluation Environments}
Our evaluation needs to be parallelizable and capable of multiple grasp evaluations on the same scene.
Since all scenes contained objects that were more exposed and thus preferred targets for grasp synthesis methods,
we designed an evaluation procedure that simulates object removal in each iteration. Given the number of graspable
objects in each scene, we evaluated the number of successful grasps out of 100 generated grasps.
Iteratively, we removed a single graspable object and repeated the process until no graspable objects remained.
This metric, while biased towards object count, standardized the evaluation and allowed for parallelization
and pre-generation of scenes while preserving relevant performance indicators.
For all our evaluations, we tested on ten scenes in which each graspable object contained at least 100
valid grasps out of the 5,000 previously generated.

\paragraph{Single-Gripper Performance}
We first validated the improvements of our method against the original
Diffusion-EDF \cite{ryu2023diffusion} algorithm.
As shown in Table \ref{table:cgn}, our method shows a notable enhancement in single-gripper grasping performance. Additionally, we compared the Diffusion-EDF architecture \cite{ryu2023diffusion} with a comparable non-equivariant U-Net–based variant, where we replaced the equiformer layers with similar
graph attention layers \cite{brody2022how}. This variant has three times as many parameters and
requires ten times the training iterations. The results indicate a performance
deficit in the non-equivariant model, demonstrating the effectiveness of equivariant methods for this problem.
Furthermore, we compared our approach to the state-of-the-art Contact-GraspNet \cite{sundermeyer2021contact} (CGN).
Our method sampled 100 grasps without post-processing, while CGN utilized the 100 most confident grasps from ten
different camera angles. In this comparison, we only included CGN with its original model weights. 
Our dataset does not provide the specific contact information required by CGN,
preventing proper training of the method on our benchmark. As such, we only used the inference
pipeline of CGN's method to predict point-wise
grasps from our RGB-D images captured in simulation.
Despite these limitations, CGN's results remain competitive. It is worth noting that CGN was originally
trained on a larger dataset comprising over 10,000 scenes, with scenes being rendered dynamically during training.
In contrast, our dataset consists of fixed rendered point clouds and contains four times fewer scenes.
This difference in dataset size and rendering approach may impact the comparative performance between our method
and CGN.

\begin{table*}[h]
\centering
\vspace{5pt}
\begin{tabular}{l|ccc}
\textbf{Method} & \textbf{Robotiq 2F-85} & \textbf{DEX-EE Hand} & \textbf{Shadow Hand (three finger)} \\ \hline
AdaGrasp (original)& 20.0 & 5.0 & 5.0 \\
AdaGrasp (retrained)& 35.0 & 15.0 & 10.0 \\
\hline
 {Without closed gripper} & 82.8 & \textbf{79.3} & 58.3 \\
 {Without batch training} & 86.1 & 75.7 & 76.5 \\
 {With equivariant FiLM} & 88.1 & 78.9 & \textbf{80.9} \\
 {Full model} & \textbf{91.3} & 75.8 & 75.7 \\
\hline
 {No Robotiq 2F-85 (zero-shot)} & 83.2 & 75.2 & \textbf{82.4} \\
 {No three finger Shadow Hand (zero-shot)} & 87.7 & 75.0 & 59.5 \\
 {No Shadow Hand (zero-shot)} & \textbf{93.0} & \textbf{78.8} & 37.6 \\
 {Robotiq 2F-85 Hand few-shot (10 shots)} & 84.8 & - & - \\
 {Shadow Hand few-shot (10 shots)} & - & - & 45.8 \\
\end{tabular}
\caption{\textbf{Grasp Success Rates (\%) across three gripper variants}.
Models trained on compatible data consisting of 450 scenes for each gripper variant (4500 scenes total). 
Results show both original and retrained model variants for AdaGrasp \cite{xu2021adagrasp}.
The full model includes the entire dataset, while zero-shot models exclude certain portions, 
highlighting zero-shot grasp transfer capabilities. Notably, the Shadow Hand dataset includes 
four distinct gripper configurations, and the corresponding partial model excludes 1350 scenes.}
\label{table:results}
\end{table*}

\begin{table*}[h]
\centering
\begin{tabular}{l|ccc}
\textbf{Method} & \textbf{Robotiq 2F-85} & \textbf{DEX-EE Hand} & \textbf{Shadow Hand (three finger)} \\ \hline
Trained single-object; Evaluated single-object & \textbf{99.1} & 97.2 & 91.7 \\
Trained clutter-object; Evaluated single-object & 97.2 & \textbf{98.6} & \textbf{93.0} \\
\hline
Trained single-object; Evaluated clutter-object & 67.6 & 40.4 & 27.7 \\
Trained clutter-object; Evaluated clutter-object & \textbf{88.1} & \textbf{78.9} & \textbf{80.9} \\
\end{tabular}
\caption{\textbf{Grasp Success Rates (\%) Single-object and Object-clutter scenes}. Our models were trained
and evaluated on both single-object and object-clutter scenes, consisting of 450 scenes (4,500 scenes in total).
The model trained on cluttered scenes outperforms the single-object model by a substantial margin on the
clutter-object benchmark.}
\label{table:single_vs_clutter}
\end{table*}

\paragraph{Multi-Gripper Performance}
We compared the multi-gripper performance of our method with AdaGrasp \cite{xu2021adagrasp}.
For this comparison, we created a separate dataset consisting of 450 scenes per gripper, each
containing 32 valid and 32 invalid top-grasps within a restricted workspace.
Each scene included up to five objects placed flat on a table because AdaGrasp does not account for
object heaps. We evaluated the methods by executing one grasp per scene over twenty
workspace-restricted scenes.
To ensure a fair comparison, we deviated from AdaGrasp's reinforcement-style data generation
and trained the method using supervised learning. Standard epoch-based learning proved
unsuccessful, prompting us to adopt a biased training procedure that prioritized samples
with significant discrepancies between predicted scores and ground truth labels. These
surprise labels were adjusted through Polyak averaging during training. TSDF volumes were
rendered from ten images per scene and gripper, matching AdaGrasp’s resolution.
Results for both the retrained and original model weights are presented in Table
\ref{table:results},
where our method demonstrates a clear advantage across all gripper types.
Additionally, we evaluated our model trained on both single-object and cluttered scenes.
The results in Table \ref{table:single_vs_clutter} clearly indicate that a model trained on single-object
scenes does not generalize effectively to cluttered environments, whereas training on
cluttered scenes allows for effective generalization to single-object scenarios.
These findings underscore the importance of training with complex grasping scenarios to achieve
robust performance in diverse environments.
Selected scenes with synthesized grasps are shown in Figure \ref{fig:collision}.
In our ablation study, we assessed the impact of key model modifications. The results indicate
that both enabling batch learning and incorporating closed gripper geometry significantly
enhance performance on the Shadow Hand and Robotiq 2F-85 grippers. Experiments with the
equivariant FiLM \cite{perez2018film} model variant also demonstrate promise. In conclusion, we
observe strong generalizability across all model variants, with performance increases for each
introduced modification, particularly evident for the complex gripper type.

\paragraph{Zero- and Few-Shot Performance}
Our zero-shot and few-shot performance evaluations are presented in Table
\ref{table:results}. To simulate a zero-shot setting, we excluded specific gripper types from the training set and trained models on these restricted subsets. All models were trained on an A100 GPU for the same duration; however, due to increased computational demands, the equivariant FiLM model variant was not employed for these tasks.
The full model, trained on data from all gripper types, significantly outperformed the baseline model across all tested grippers. Zero-shot models exhibited reduced performance when evaluated on unseen gripper types, underscoring the challenges of zero-shot grasp transfer.
In the few-shot scenario, we fine-tuned the zero-shot models using ten scenes of the missing gripper type, each containing only one grasp, and trained for a few epochs. This fine-tuning led to noticeable improvements in performance, suggesting that our approach is beneficial when only a limited number of grasps can be manually gathered for an unknown gripper type.
\begin{table}[h]
\centering
\begin{tabular}{l|ccc}
\textbf{Experiment} & \textbf{Robotiq 2F-140} & \textbf{Schunk WSG-32} \\ \hline
{Zero-shot bin picking} & 82.0 & 88.0 \\
\end{tabular}
\caption{\textbf{Grasp Success Rates (\%) in Real-World Bin Picking Experiments}. The table summarizes the grasp
success rates calculated from 50 executed grasps achieved in zero-shot bin picking tasks using the Robotiq 2F-140,
Schunk WSG-32 grippers. All experiments were conducted on objects and grippers not included in the training dataset,
demonstrating the model's generalization capabilities.}
\label{table:real_world_results}
\end{table}

\begin{figure}
    \centering
    % First minipage for the tall image (a)
    \begin{minipage}[b]{0.35\linewidth}
        \centering
        \vspace{5pt}
        \includegraphics[width=\linewidth]{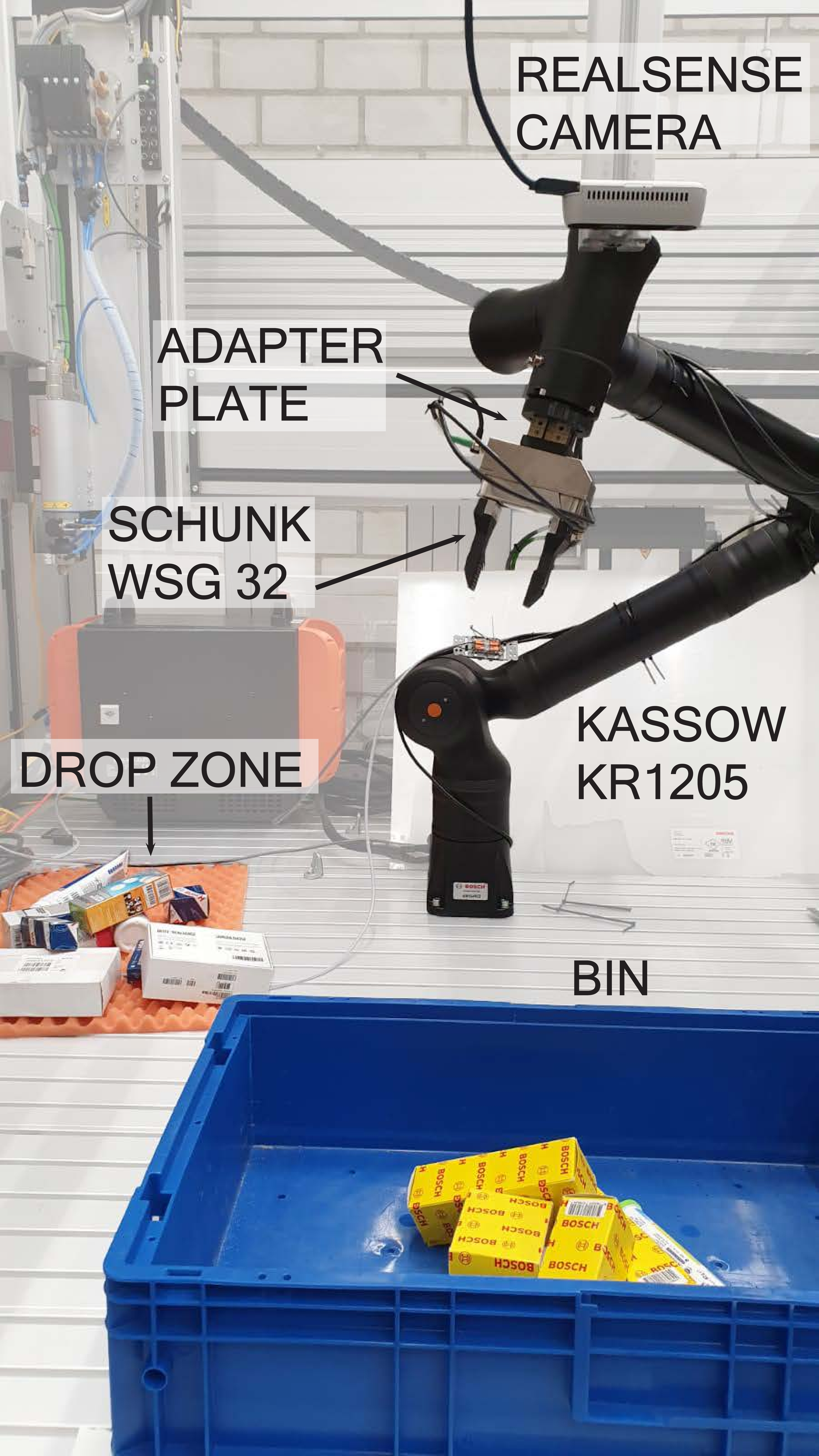}
        \smallskip\\
        \textbf{(a)}
    \end{minipage}%
    \hfill
    % Second minipage for the two stacked images (b)
    \begin{minipage}[b]{0.4\linewidth}
        \centering
        \includegraphics[width=\linewidth]{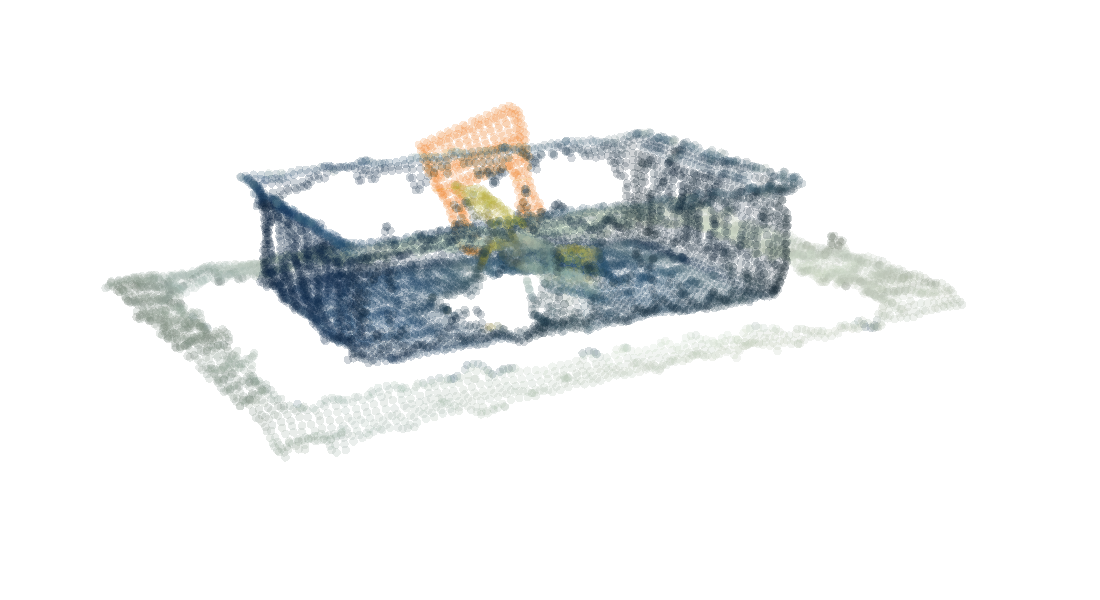}\\[0.5cm]
        \includegraphics[width=\linewidth]{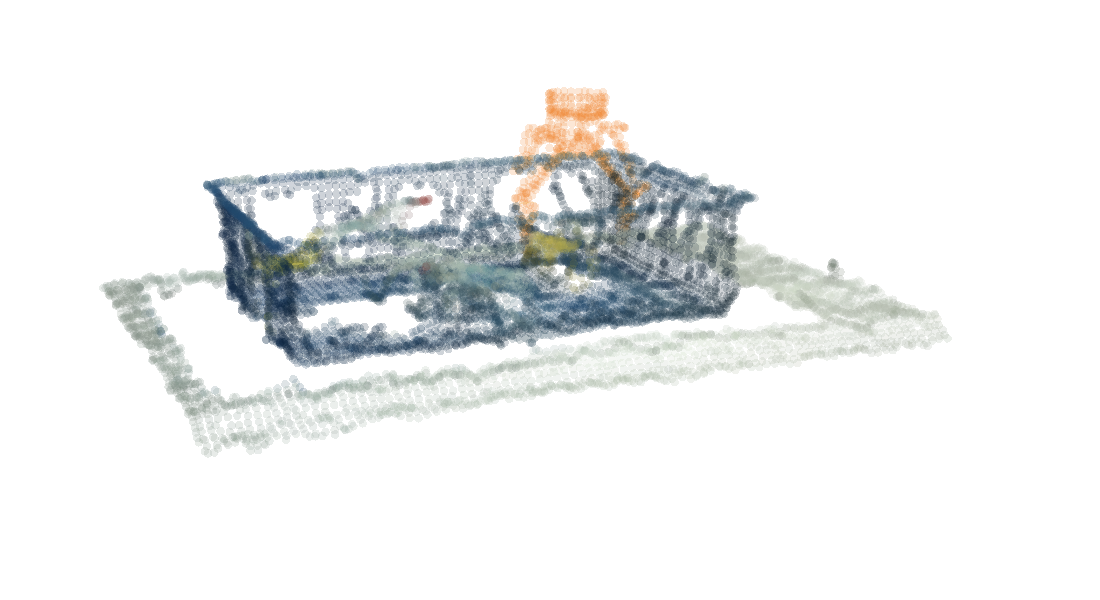}
        \smallskip\\
        \textbf{(b)}
    \end{minipage}
    \caption{\textbf{Real-world setup.} (a) The Kassow KR 1205 robot with 7 axes,
    equipped with a Schunk WSG-32 gripper and a Schunk tool changer for gripper exchange.
    A top-down RealSense camera is used to capture the scene's point cloud.
    (b) Visualizations of point clouds including generated grasps for the WSG-32 and Robotiq 2F-85 grippers.
    Note that the grippers and objects are not part of the training dataset.}
    \label{fig:kassow-explained}
\end{figure}

\paragraph{Real-World Laboratory Experiments}
Figure \ref{fig:kassow-explained} illustrates our hardware setup for testing zero-shot and sim-to-real
transfer under real-world conditions. We utilized the 7-axis Kassow KR 1205 robot, equipped
with a custom Schunk gripper exchanger system for quick gripper replacement.
For these experiments, we trained a separate model on bin picking scenes, filtering out grasps with an angle
greater than 25 degrees from the gravity z-direction.
Our grasp synthesis inference pipeline takes approximately 35 seconds to generate a batch of
20 grasps on an A2000 GPU. We did not apply any post-processing to the generated grasps and
consistently used the first grasp from each batch for evaluation.
During execution, we positioned the gripper at the proposed pre-grasp pose with a z-offset, then moved
the gripper to the predicted height and closed the gripper.
Trajectory collision avoidance was implemented only with respect to the test cell and non-movable objects;
objects inside the bin were not considered. During evaluation, we tested the Robotiq 2F-140 and Schunk WSG-32 grippers.
A RealSense camera was mounted with a top-down view of the bin. To obtain point clouds of the gripper types,
we used synthetic rendering of the CAD models of these grippers. Neither the objects nor the grippers used in the tests
were not included in our training dataset. Among the objects in the bin, we always included ten graspable
and two obstacle objects and performed a bin clearing procedure. The test object set includes everyday 
household items, packaging boxes, and non-empty bottles, with shapes ranging from boxes and tubes
to more complex forms, such as hollow objects like tape rolls.
We conducted five series of bin picking tasks for each gripper variant,
calculating the average grasp success rate from 50 grasps,
as presented in Table \ref{table:real_world_results}. All evaluated models were
trained solely on simulation data. While the demonstrated performance is noteworthy,
we observed a few common failure cases. More than five of the failed grasps were not related
to the method and were mainly due to either failed collision avoidance in the trajectory or
grasp execution precision. Specifically, smaller objects were sometimes misclassified as
being ungraspable. In such cases, when no other graspable object was present,
the gripper attempted to grasp the bin.

\section{Limitations and Future Work}
\label{sec:future}

Our work aims to develop architectures that incorporate various input features—such as color, texture,
physical properties like friction and applied forces, and synthetic features from visual foundation models
\cite{ravi2024sam2}. While we primarily rely on visual features in this study, our initial encoding
preprocesses data into a standard format while preserving spatial information, facilitating future inclusion of
additional features. We represent joint configurations in a kinematic-agnostic manner by encoding the gripper as
a featurized point cloud, excluding internal DoF in this initial step toward a generalist grasp
synthesis method. This simplification helps demonstrate our approach but requires further validation and extension,
particularly as we aim to incorporate variable internal DoF and kinematic features in future work.
Our heuristic gripper encoder uses renderings of the gripper in both open and closed states, simplifying kinematic
encoding and providing information about ambiguous states, especially for high-DoF grippers.
While adequate for industrial two- and three-finger grippers like the Schunk variants, this design will need
revision as we progress.
We encountered disproportionate memory and computational bottlenecks due to unoptimized equivariant libraries.
We anticipate resolving these issues with more optimized machine learning frameworks like JAX \cite{jax2018github}
and ongoing hardware advancements. In this work, we deprioritized computational optimization, deferring performance
improvements to future iterations.
Our original grasp sampling strategy treats all gripper variants as parallel-jaw grippers, enabling large-scale
parallelization but causing loss of gripper-specific grasps and lower success rates for non-parallel grippers
like the Shadow Hand. Certain gripper-object combinations remain infeasible and are absent from our dataset,
limiting the performance of gripper-agnostic algorithms in our benchmarks.
Nevertheless, the few-shot cases demonstrate a viable method deployable in common industrial settings—such
as bin picking and simple manipulation tasks—with the ability to quickly adapt to novel gripper designs. We believe that further expanding the gripper and object dataset will enhance the zero-shot capabilities of our approach.

\section{Conclusion} 
In this work, we introduced a framework for multi-embodiment grasp synthesis using equivariant
diffusion models. Our approach achieves state-of-the-art performance in single-gripper scenarios
and demonstrates robust generalization across multiple gripper types, significantly surpassing
state-of-the-art methods in grasp quality and success rates.

\bibliographystyle{IEEEtran}
\bibliography{references}

\end{document}